\documentclass[10pt,twocolumn,letterpaper]{article}

\usepackage[pagenumbers]{cvpr} 

\usepackage[dvipsnames]{xcolor}

\definecolor{cvprblue}{rgb}{0.21,0.49,0.74}
\usepackage[pagebackref,breaklinks,colorlinks,citecolor=cvprblue]{hyperref}

\def\paperID{*****} 
\def\confName{CVPR}
\def\confYear{2024}

\title{Zero-shot Compound Expression Recognition with Visual Language Model at the 6th ABAW Challenge}

\author{Jiahe Wang\\
University of Science and Technology of China\\
Hefei, China\\
{\tt\small wangjiahe317@mail.ustc.edu.cn}
\and
Jiale Huang\\
University of Science and Technology of China\\
Hefei, China\\
{\tt\small huangjiale@mail.ustc.edu.cn}
\and
Bingzhao Cai\\
University of Science and Technology of China\\
Hefei, China\\
{\tt\small cbz\_2020@mail.ustc.edu.cn}
\and
Yifan Cao\\
University of Science and Technology of China\\
Hefei, China\\
{\tt\small yyff@mail.ustc.edu.cn}
\and
Xin Yun\\
University of Science and Technology of China\\
Hefei, China\\
{\tt\small yx9329@mail.ustc.edu.cn}
\and
Shangfei Wang\\
University of Science and Technology of China\\
Hefei, China\\
{\tt\small sfwang@ustc.edu.cn}
}

\begin{document}
\maketitle
\begin{abstract}
Conventional approaches to facial expression recognition primarily focus on the classification of six basic facial expressions.  Nevertheless, real-world situations present a wider range of complex compound expressions that consist of combinations of these basics ones due to limited availability of comprehensive training datasets.  The 6th Workshop and Competition on Affective Behavior Analysis in-the-wild (ABAW) offered unlabeled datasets containing compound expressions.  In this study, we propose a zero-shot approach for recognizing compound expressions by leveraging a pretrained visual language model integrated with some traditional CNN networks. \textit{The code and models is available at \url{https://github.com/Jialehuang-nb/Z-CER}.}
\end{abstract}    
\section{Introduction}
\label{sec:intro}

The recognition of facial expressions is an essential component in the field of artificial intelligence, as it enables computers to effectively convey human emotional information\cite{TARNOWSKI20171175}, complementing tone as a crucial means of emotional communication in real-life scenarios.

Traditional facial expression recognition is limited to the classification of six basic facial expressions\cite{doi:10.1073/pnas.1322355111}, namely anger, happiness, sadness, surprise, disgust and fear. However, in real-life situations, human emotions exhibit much greater complexity than these predefined categories. The introduction of Compound expressions opens up a new avenue for facial expression recognition research and has the potential to elevate the fields of computer vision and artificial intelligence to a higher level.  Compound expressions are typically captured from uncontrolled environments without any specific conditions imposed.  Most publicly available datasets on naturalistic facial expressions only encompass basic expressions, while there is a scarcity of training data for datasets containing Compound expressions.

The participants in the Challenge of the 6th Workshop and Competition on Affective Behavior Analysis in-the-wild (ABAW) for Compound Expression (CE) Recognition were provided with a portion of the C-EXPR-DB\cite{kollias20246th,kollias2023abaw2,kollias2023multi,kollias2023abaw,kollias2022abaw,kollias2021analysing,kollias2020analysing,kollias2021distribution,kollias2021affect,kollias2019expression,kollias2019face,kollias2019deep,zafeiriou2017aff} database, consisting of 56 unannotated videos. C-EXPR-DB is an audiovisual (A/V) in-the-wild database comprising a total of 400 videos, containing approximately 200K frames, each annotated with respect to 12 compound expressions. For this Challenge, only seven compound expressions will be considered: Fearfully Surprised, Happily Surprised, Sadly Surprised, Disgustedly Surprised, Angrily Surprised, Sadly Fearful and Sadly Angry.

The rapid advancements in large-scale visual language pre-trained models have led us to opt for the utilization of the Claude3\cite{claude3} model in order to enhance our compound expression recognition capabilities.
\section{Method}
\begin{figure*}
    \centering
    \includegraphics[width=\textwidth]{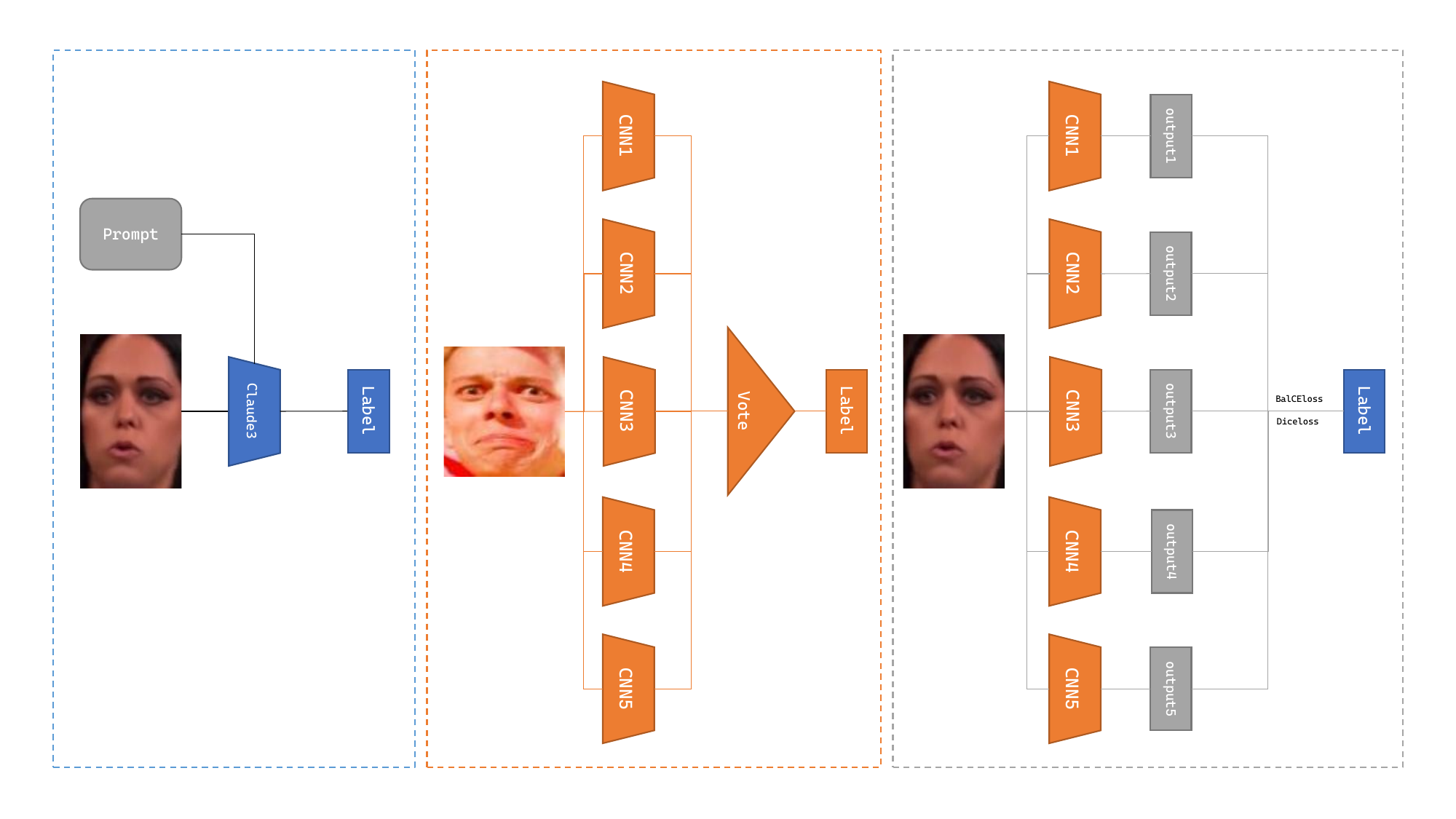}
    \caption{Overview of our proposed framework}
    \label{fig:framework}
\end{figure*}
The specific flow of our approach is depicted in Figure \ref{fig:framework}.  Initially, we annotate a subset of the unlabeled data using a pre-trained visual language model.  Subsequently, we train five CNN classification networks with ground truth label data.  Finally, we fine-tune the CNNs utilizing the labels generated by the visual language model.

The component enclosed in the blue box represents the pre-trained visual language model, while the component within the orange box corresponds to the CNNs. Lastly, the segment contained in the gray box signifies the fine-tuning process.
\subsection{Data Process}
We use raw unlabeled C-EXPR-DB\cite{kollias20246th,kollias2023abaw2,kollias2023multi,kollias2023abaw,kollias2022abaw,kollias2021analysing,kollias2020analysing,kollias2021distribution,kollias2021affect,kollias2019expression,kollias2019face,kollias2019deep,zafeiriou2017aff} as our training and testing database, which has 56 different videos, including one of 7 different compound expressions per frame. Following previous paper\cite{kollias2023multi} , we firstly divide all the videos into individual frame, then Retinaface\cite{deng2020retinaface} is used to detect, crop and align the faces in these frames. 

However, in the raw C-EXPR-DB, There are frames with multiple faces or frames without faces. For multiple faces cases, to avoid contradictions of emotions among different individuals, we only select the face of the “protagonist”, who has the biggest face region, as the basis for this frame. For no face cases, since emotions have continuity, we copy the face in closest frames to the current frame as the basis. After all, each frame has only one face for discrimination.

\subsection{Visual Language Model}
\begin{table}[]
    \centering
    \begin{tabular}{|p{0.9\columnwidth}|}
    \hline
       Which of the following expressions is the expression in the image: 'Happily Surprised', 'Sadly Fearful', 'Sadly Angry', 'Sadly Surprised', 'Fearfully Surprised', 'Angrily Surprised', 'Disgustedly Surprised'? No explanation needed, just give me the name of a category    \\
       \hline 
       Which of these expressions in the picture is like 'Happily Surprised', 'Sadly Fearful', 'Sadly Angry', 'Sadly Surprised', 'Fearfully Surprised', 'Angrily Surprised' or 'Disgustedly Surprised'? No need for an explanation, just give me the name of a category.   \\
       \hline
    \end{tabular}
    \caption{Prompts}
    \label{tab:prompts}
\end{table}
The API provided by Claude official is utilized to access the Claude3 model, specifically employing the text presented in Table \ref{tab:prompts} as the prompt words. Upon receiving the input prompt text and image, the model will generate a paragraph of textual description elucidating the facial behavior depicted in the figure. Keyword extraction is employed to acquire high-confidence labels serving as pseudo-labels for the input image.
\subsection{CNNs}
In the CNNs section, we employed a total of five distinct convolution neural networks (CNNs), namely mobilenetV2\cite{sandler2019mobilenetv2}, resnet152\cite{he2015deep}, densenet121\cite{huang2018densely}, resnet18\cite{he2015deep}, and densenet201\cite{huang2018densely}.  The corresponding parameter numbers are presented in Table \ref{tab:num_params}.  Initially, each classification network was trained individually followed by aggregating predictions through a majority voting mechanism to obtain the final predicted label class. During training, we utilized BalCEloss\cite{xu2023learning} and MultiDiceLoss as loss functions.  

BalCEloss is a recently proposed classification loss function specifically designed for imbalanced data with long-tail distribution (Eq. \ref{eq:1}). 
\begin{equation}
\begin{aligned}
& \mathcal{L}_{\text {Bal-CE }}\left(\mathcal{M}\left(\mathbf{x} \mid \theta_f, \theta_w\right), \mathbf{y}_i\right)=-\log \left(p\left(\mathbf{y}_i \mid \mathbf{x} ; \theta_f, \theta_w\right)\right) \\
& =-\log \left[\frac{n_{\mathbf{y}_i} e^{z_{\mathbf{y}_i}}}{\sum_{\mathbf{y}_j \in \mathcal{Y}} n_{\mathbf{y}_j} e^{z_{\mathbf{y}_j}}}\right] \\
& =\log \left[1+\sum_{\mathbf{y}_j \neq \mathbf{y}_i} e^{\log n_{\mathbf{y}_j}-\log n_{\mathbf{y}_i}} \cdot e^{\mathbf{z}_{\mathbf{y}_j}-\mathbf{z}_{\mathbf{y}_i}}\right]
\label{eq:1}
\end{aligned}
\end{equation}

On the other hand, DiceLoss optimizes the similarity between two sets and can be considered as directly optimizing F1 Score due to its similar form (Eq. \ref{eq:2}).
\begin{equation}
\mathcal{L}_\text{ DiceLoss }=1-\frac{2|\hat Y \bigcap Y|}{|\hat Y|+|Y|}
\label{eq:2}
\end{equation}
The MultiDiceLoss method transforms multi-class labels into multiple binary labels through one-hot encoding. Each channel can be regarded as a binary classification problem, thus the loss can be obtained by calculating the DiceLoss for each class in a binary classification manner and subsequently averaging the loss values across all channels to derive the final loss.

The total loss is defined as:
\begin{equation}
\mathcal{L}=\mathcal{L}_{\text {Bal-CE }} + \lambda \mathcal{L}_\text{ DiceLoss }
\end{equation}
Where $\lambda$ is a weight factor to balance the impact of the two loss on the final result. We set $\lambda$ to 1.5 in our experiments.
\begin{table}[]
    \centering
    \begin{tabular}{c|c}
     Model    & Num Params \\
     \hline
     mobilenetV2  &  3504872 \\
     resnet152  & 	60192808  \\
     densenet121 & 	7978856   \\
     resnet18   &  	11689512  \\
     densenet201 & 	20013928  \\
    \end{tabular}
    \caption{Params Num}
    \label{tab:num_params}
\end{table}

\section{Experiments}
\subsection{Experimental database}
We use two databases for our experiments, namely RAF-DB\cite{li2017reliable,li2019reliable} and the C-EXPR-DB\cite{kollias20246th,kollias2023abaw2,kollias2023multi,kollias2023abaw,kollias2022abaw,kollias2021analysing,kollias2020analysing,kollias2021distribution,kollias2021affect,kollias2019expression,kollias2019face,kollias2019deep,zafeiriou2017aff} provided by the 6th Workshop and Competition on Affective Behavior Analysis in-the-wild (ABAW).

The Real-world Affective Faces Database (RAF-DB) is a comprehensive facial expression database consisting of approximately 30,000 diverse facial images sourced from the Internet. Each image has been meticulously labeled by around 40 annotators through crowdsourcing annotation. The database encompasses a wide range of variations in subjects' age, gender, ethnicity, head poses, lighting conditions, occlusions (such as glasses, facial hair or self-occlusion), and post-processing operations (including various filters and special effects). RAF-DB offers extensive diversities, abundant quantities, and rich annotations including: 29,672 real-world images; a 7-dimensional expression distribution vector for each image; two distinct subsets - a single-label subset comprising seven classes of basic emotions and a two-tab subset encompassing twelve classes of compound emotions; accurate landmark locations for five points; automatic landmark locations for thirty-seven points; bounding box information; race classification; age range estimation; and gender attribute annotations per image. We use seven of the total twelve classes of compound emotions, ensuring that these specific categories aligned with the requirements of the competition.

C-EXPR-DB is an audiovisual (A/V) in-the-wild database comprising a total of 400 videos, containing approximately 200K frames, each annotated with respect to twelve compound expressions. For this Challenge, only seven compound expressions will be considered: Fearfully Surprised, Happily Surprised, Sadly Surprised, Disgustedly Surprised, Angrily Surprised, Sadly Fearful and Sadly Angry.

\subsection{Implementation Details}
We utilize the pretrained weights provided by torchvision\cite{torchvision} as initial values for our 5 CNNs. The classification layer of each CNN is constructed using two linear layers and a nonlinear activation layer, with an output dimension of seven dimensions. The overall framework is implemented using PyTorch\cite{paszke2019pytorch}. The Adam optimizer\cite{kingma2017adam} is employed to optimize each network. The batch size and learning rate are set to 64 and 0.0001, respectively. 

\subsection{Evaluation Metrics}
According to the performance assessment rules of the competition\cite{kollias20246th}, we evaluate the performance of compound expressions recognition by the average F1 Score across all 7 compound expressions.  Therefore, the evaluation criterion is:

\begin{equation}
\mathcal{P}_{C E}=\frac{\sum_{\text {expr }} F_1^{\text {expr }}}{7}
\end{equation}
{
    \small
    \bibliographystyle{ieeenat_fullname}
    \bibliography{main}

\begin{thebibliography}{26}
\providecommand{\natexlab}[1]{#1}
\providecommand{\url}[1]{\texttt{#1}}
\expandafter\ifx\csname urlstyle\endcsname\relax
  \providecommand{\doi}[1]{doi: #1}\else
  \providecommand{\doi}{doi: \begingroup \urlstyle{rm}\Url}\fi

\bibitem[Anthropic()]{claude3}
Anthropic.
\newblock Claude.
\newblock \url{https://www.anthropic.com/claude}.
\newblock 2024/3/15.

\bibitem[Deng et~al.(2020)Deng, Guo, Ververas, Kotsia, and
  Zafeiriou]{deng2020retinaface}
Jiankang Deng, Jia Guo, Evangelos Ververas, Irene Kotsia, and Stefanos
  Zafeiriou.
\newblock Retinaface: Single-shot multi-level face localisation in the wild.
\newblock In \emph{Proceedings of the IEEE/CVF conference on computer vision
  and pattern recognition}, pages 5203--5212, 2020.

\bibitem[Du et~al.(2014)Du, Tao, and Martinez]{doi:10.1073/pnas.1322355111}
Shichuan Du, Yong Tao, and Aleix~M. Martinez.
\newblock Compound facial expressions of emotion.
\newblock \emph{Proceedings of the National Academy of Sciences}, 111\penalty0
  (15):\penalty0 E1454--E1462, 2014.

\bibitem[He et~al.(2015)He, Zhang, Ren, and Sun]{he2015deep}
Kaiming He, Xiangyu Zhang, Shaoqing Ren, and Jian Sun.
\newblock Deep residual learning for image recognition, 2015.

\bibitem[Huang et~al.(2018)Huang, Liu, van~der Maaten, and
  Weinberger]{huang2018densely}
Gao Huang, Zhuang Liu, Laurens van~der Maaten, and Kilian~Q. Weinberger.
\newblock Densely connected convolutional networks, 2018.

\bibitem[Kingma and Ba(2017)]{kingma2017adam}
Diederik~P. Kingma and Jimmy Ba.
\newblock Adam: A method for stochastic optimization, 2017.

\bibitem[Kollias(2022)]{kollias2022abaw}
Dimitrios Kollias.
\newblock Abaw: Valence-arousal estimation, expression recognition, action unit
  detection \& multi-task learning challenges.
\newblock In \emph{Proceedings of the IEEE/CVF Conference on Computer Vision
  and Pattern Recognition}, pages 2328--2336, 2022.

\bibitem[Kollias(2023{\natexlab{a}})]{kollias2023abaw}
Dimitrios Kollias.
\newblock Abaw: Learning from synthetic data \& multi-task learning challenges.
\newblock In \emph{European Conference on Computer Vision}, pages 157--172.
  Springer, 2023{\natexlab{a}}.

\bibitem[Kollias(2023{\natexlab{b}})]{kollias2023multi}
Dimitrios Kollias.
\newblock Multi-label compound expression recognition: C-expr database \&
  network.
\newblock In \emph{Proceedings of the IEEE/CVF Conference on Computer Vision
  and Pattern Recognition}, pages 5589--5598, 2023{\natexlab{b}}.

\bibitem[Kollias and Zafeiriou(2019)]{kollias2019expression}
Dimitrios Kollias and Stefanos Zafeiriou.
\newblock Expression, affect, action unit recognition: Aff-wild2, multi-task
  learning and arcface.
\newblock \emph{arXiv preprint arXiv:1910.04855}, 2019.

\bibitem[Kollias and Zafeiriou(2021{\natexlab{a}})]{kollias2021affect}
Dimitrios Kollias and Stefanos Zafeiriou.
\newblock Affect analysis in-the-wild: Valence-arousal, expressions, action
  units and a unified framework.
\newblock \emph{arXiv preprint arXiv:2103.15792}, 2021{\natexlab{a}}.

\bibitem[Kollias and Zafeiriou(2021{\natexlab{b}})]{kollias2021analysing}
Dimitrios Kollias and Stefanos Zafeiriou.
\newblock Analysing affective behavior in the second abaw2 competition.
\newblock In \emph{Proceedings of the IEEE/CVF International Conference on
  Computer Vision}, pages 3652--3660, 2021{\natexlab{b}}.

\bibitem[Kollias et~al.()Kollias, Schulc, Hajiyev, and
  Zafeiriou]{kollias2020analysing}
D Kollias, A Schulc, E Hajiyev, and S Zafeiriou.
\newblock Analysing affective behavior in the first abaw 2020 competition.
\newblock In \emph{2020 15th IEEE International Conference on Automatic Face
  and Gesture Recognition (FG 2020)(FG)}, pages 794--800.

\bibitem[Kollias et~al.(2019{\natexlab{a}})Kollias, Sharmanska, and
  Zafeiriou]{kollias2019face}
Dimitrios Kollias, Viktoriia Sharmanska, and Stefanos Zafeiriou.
\newblock Face behavior a la carte: Expressions, affect and action units in a
  single network.
\newblock \emph{arXiv preprint arXiv:1910.11111}, 2019{\natexlab{a}}.

\bibitem[Kollias et~al.(2019{\natexlab{b}})Kollias, Tzirakis, Nicolaou,
  Papaioannou, Zhao, Schuller, Kotsia, and Zafeiriou]{kollias2019deep}
Dimitrios Kollias, Panagiotis Tzirakis, Mihalis~A Nicolaou, Athanasios
  Papaioannou, Guoying Zhao, Bj{\"o}rn Schuller, Irene Kotsia, and Stefanos
  Zafeiriou.
\newblock Deep affect prediction in-the-wild: Aff-wild database and challenge,
  deep architectures, and beyond.
\newblock \emph{International Journal of Computer Vision}, pages 1--23,
  2019{\natexlab{b}}.

\bibitem[Kollias et~al.(2021)Kollias, Sharmanska, and
  Zafeiriou]{kollias2021distribution}
Dimitrios Kollias, Viktoriia Sharmanska, and Stefanos Zafeiriou.
\newblock Distribution matching for heterogeneous multi-task learning: a
  large-scale face study.
\newblock \emph{arXiv preprint arXiv:2105.03790}, 2021.

\bibitem[Kollias et~al.(2023)Kollias, Tzirakis, Baird, Cowen, and
  Zafeiriou]{kollias2023abaw2}
Dimitrios Kollias, Panagiotis Tzirakis, Alice Baird, Alan Cowen, and Stefanos
  Zafeiriou.
\newblock Abaw: Valence-arousal estimation, expression recognition, action unit
  detection \& emotional reaction intensity estimation challenges.
\newblock In \emph{Proceedings of the IEEE/CVF Conference on Computer Vision
  and Pattern Recognition}, pages 5888--5897, 2023.

\bibitem[Kollias et~al.(2024)Kollias, Tzirakis, Cowen, Zafeiriou, Kotsia,
  Baird, Gagne, Shao, and Hu]{kollias20246th}
Dimitrios Kollias, Panagiotis Tzirakis, Alan Cowen, Stefanos Zafeiriou, Irene
  Kotsia, Alice Baird, Chris Gagne, Chunchang Shao, and Guanyu Hu.
\newblock The 6th affective behavior analysis in-the-wild (abaw) competition,
  2024.

\bibitem[Li and Deng(2019)]{li2019reliable}
Shan Li and Weihong Deng.
\newblock Reliable crowdsourcing and deep locality-preserving learning for
  unconstrained facial expression recognition.
\newblock \emph{IEEE Transactions on Image Processing}, 28\penalty0
  (1):\penalty0 356--370, 2019.

\bibitem[Li et~al.(2017)Li, Deng, and Du]{li2017reliable}
Shan Li, Weihong Deng, and JunPing Du.
\newblock Reliable crowdsourcing and deep locality-preserving learning for
  expression recognition in the wild.
\newblock In \emph{2017 IEEE Conference on Computer Vision and Pattern
  Recognition (CVPR)}, pages 2584--2593. IEEE, 2017.

\bibitem[Paszke et~al.(2019)Paszke, Gross, Massa, Lerer, Bradbury, Chanan,
  Killeen, Lin, Gimelshein, Antiga, Desmaison, Köpf, Yang, DeVito, Raison,
  Tejani, Chilamkurthy, Steiner, Fang, Bai, and Chintala]{paszke2019pytorch}
Adam Paszke, Sam Gross, Francisco Massa, Adam Lerer, James Bradbury, Gregory
  Chanan, Trevor Killeen, Zeming Lin, Natalia Gimelshein, Luca Antiga, Alban
  Desmaison, Andreas Köpf, Edward Yang, Zach DeVito, Martin Raison, Alykhan
  Tejani, Sasank Chilamkurthy, Benoit Steiner, Lu Fang, Junjie Bai, and Soumith
  Chintala.
\newblock Pytorch: An imperative style, high-performance deep learning library,
  2019.

\bibitem[pytorch.org()]{torchvision}
pytorch.org.
\newblock Torchvision.models.
\newblock \url{https://pytorch.org/vision/stable/models.html}.
\newblock 2024/3/17.

\bibitem[Sandler et~al.(2019)Sandler, Howard, Zhu, Zhmoginov, and
  Chen]{sandler2019mobilenetv2}
Mark Sandler, Andrew Howard, Menglong Zhu, Andrey Zhmoginov, and Liang-Chieh
  Chen.
\newblock Mobilenetv2: Inverted residuals and linear bottlenecks, 2019.

\bibitem[Tarnowski et~al.(2017)Tarnowski, Kołodziej, Majkowski, and
  Rak]{TARNOWSKI20171175}
Paweł Tarnowski, Marcin Kołodziej, Andrzej Majkowski, and Remigiusz~J. Rak.
\newblock Emotion recognition using facial expressions.
\newblock \emph{Procedia Computer Science}, 108:\penalty0 1175--1184, 2017.
\newblock International Conference on Computational Science, ICCS 2017, 12-14
  June 2017, Zurich, Switzerland.

\bibitem[Xu et~al.(2023)Xu, Liu, Yang, Chai, and Yuan]{xu2023learning}
Zhengzhuo Xu, Ruikang Liu, Shuo Yang, Zenghao Chai, and Chun Yuan.
\newblock Learning imbalanced data with vision transformers, 2023.

\bibitem[Zafeiriou et~al.(2017)Zafeiriou, Kollias, Nicolaou, Papaioannou, Zhao,
  and Kotsia]{zafeiriou2017aff}
Stefanos Zafeiriou, Dimitrios Kollias, Mihalis~A Nicolaou, Athanasios
  Papaioannou, Guoying Zhao, and Irene Kotsia.
\newblock Aff-wild: Valence and arousal ‘in-the-wild’challenge.
\newblock In \emph{Computer Vision and Pattern Recognition Workshops (CVPRW),
  2017 IEEE Conference on}, pages 1980--1987. IEEE, 2017.

\end{thebibliography}
}


\end{document}